\documentclass[10pt,twocolumn,letterpaper]{article}

\usepackage[T1]{fontenc}
\usepackage{times}
\usepackage[margin=0.85in]{geometry}
\usepackage{graphicx}
\usepackage{amsmath,amssymb}
\usepackage{booktabs}
\usepackage{multirow}
\usepackage{algorithm}
\usepackage{algpseudocode}
\usepackage{caption}
\usepackage[colorlinks=true,citecolor=blue,linkcolor=blue,urlcolor=blue]{hyperref}
\usepackage{xcolor}
\usepackage{balance}
\usepackage{enumitem}
\setlist{nosep,leftmargin=*}

\title{\Large \bf Physics-Grounded Fluid Video Generation with a Simulation Dataset and Dual-Stream Optical-Flow Supervision}

\author{
\small
Ruijie Su$^{1,*}$ \quad Yuanzhi Liang$^{2,*}$ \quad Xiaohua Xie$^{1,3, \dagger}$ \quad Jianhuang Lai$^{1,3}$\\[2pt]
\small
$^{1}$School of Computer Science and Engineering, Sun Yat-sen University, Guangzhou, China\\
\small
$^{2}$Institute of Artificial Intelligence, China Telecom (TeleAI), China\\
\small
$^{3}$Guangdong Province Key Laboratory of Information Security Technology, Sun Yat-sen University, Guangzhou, China
}

\date{}

\begin{document}

\twocolumn[
\maketitle
\begin{center}
\begin{minipage}{0.92\textwidth}
\centering
\small\bf Abstract
\end{minipage}
\end{center}
\vspace{2pt}
\begin{minipage}{\textwidth}
\small
Video diffusion models generate visually compelling content but routinely violate elementary physics when the subject involves fluids: liquid columns break apart in mid-air, container water levels fail to rise as liquid is poured in, and splashes disperse without regard to momentum or gravity. We attribute this gap to the fact that large-scale video-text corpora contain almost no explicit motion supervision, so models learn to imitate fluid appearance rather than dynamics. We address this with two contributions. First, we build a physics-simulation fluid dataset combining $1{,}638$ MPM-simulated pouring/sloshing videos with $2{,}320$ keyword-filtered real pouring videos mined from stock footage, plus two held-out test sets: a $1{,}515$-video real-video benchmark and an 18-prompt text-to-first-frame generalization benchmark. Second, we introduce a dual-stream image-to-video architecture built on a pretrained diffusion-transformer video generator. It augments the standard RGB decoder with a lightweight Optical-Flow Decoder branch trained with explicit end-point-error and smoothness losses, fused into the RGB stream via zero-initialized convolutions so the pretrained backbone starts undisturbed. Only the two decoders are updated; the encoder, temporal transformer, and text encoder remain frozen. Across two model scales ($1.3$B and $14$B) and two test sets, our method improves VideoPhy-2 Physical-Commonsense and Video-Quality scores over the frozen backbone by up to $8.75$ and $4.65$ points, outperforms a leading open competitor, and is preferred by human raters in a blind study. A direct optical-flow read-out evaluation further shows an end-point error as low as $0.54$ pixels in-distribution, confirming the model has internalized a coherent motion prior rather than merely improving surface appearance.
\end{minipage}
\vspace{8pt}
]

\begingroup
  \renewcommand{\footnoterule}{%
    \noindent\rule{0.48\textwidth}{0.4pt}\par
  }%
  \let\thefootnote\relax
  \footnotetext{%
    {\footnotesize\raggedright
      $^{*}$Equal contribution. $^{\dagger}$Corresponding author. \par
      Email: 472171770@qq.com\par
    }%
  }%
\endgroup

\section{Introduction}
\label{sec:intro}

Recent video diffusion and diffusion-transformer models~\cite{wan2025,cogvideox2024,videodiffusion2022,makeavideo2022} have made rapid progress in producing temporally coherent, high-resolution video from a single image or text prompt. Yet a persistent failure mode remains largely unsolved: these models do not obey the physical laws that govern the objects they depict. Fluids are an especially demanding case. Unlike rigid or piecewise-rigid objects, fluids exhibit continuous topology change, momentum transfer, and boundary-dependent splashing, all of which are difficult to infer from appearance statistics alone. In practice, state-of-the-art video generators routinely produce liquid columns that emerge from a cup wall rather than its rim, container water levels that do not rise as liquid is poured in (violating mass conservation), and splash patterns that are decoupled from the direction of impact.

We argue that this failure stems less from a fundamental architectural limitation than from a \emph{data and supervision} gap. Internet-scale video--text corpora rarely contain paired, physically-annotated fluid motion, so diffusion models trained purely with a photometric reconstruction objective have no direct incentive to learn momentum-consistent dynamics; they instead learn shortcuts that reproduce plausible-looking texture and color transitions. This motivates two complementary questions that this paper addresses: (1) how can we construct training data that isolates and exhaustively covers fluid dynamics under controlled physical parameters, and (2) how can we inject an explicit motion-level training signal into a pretrained video generator without discarding its strong prior for appearance and text alignment?

For the first question, we build a physics-simulation fluid dataset. Using a particle-based multiphase simulation framework, we procedurally generate two canonical fluid behaviors --- pouring (a container tips over a fixed pivot and empties onto a surface) and sloshing (a fluid volume is perturbed by an external horizontal force inside a fixed boundary) --- while systematically sweeping container geometry and physically meaningful parameters such as tilt speed, tilt angle, drop height, and applied force. This yields $1{,}638$ simulated videos with exact control over the underlying physics. We combine this simulated data with $2{,}320$ real pouring videos mined from a large stock-footage corpus using a fluid-noun/pouring-verb keyword filter, so that the model retains exposure to real-world visual statistics.

For the second question, we introduce a \textbf{dual-stream} fine-tuning architecture. Starting from a pretrained image-to-video diffusion transformer, we keep the visual encoder, the diffusion transformer backbone, and the text encoder entirely frozen, and add a second, lightweight decoder branch --- an \emph{Optical-Flow Decoder} --- that is trained jointly with the existing RGB decoder to directly regress the per-pixel motion field of the generated video. The two decoders share a similar architecture and are fused through zero-initialized convolutions, so that gradients from the new flow objective can influence the shared latent-to-pixel mapping without destabilizing the pretrained RGB generation quality at the start of training. Because the flow branch is supervised with pseudo ground-truth optical flow extracted from both the simulated and real training videos, the model is explicitly forced to produce a self-consistent, pixel-level motion field, rather than only a photometrically plausible final frame.

In summary, our contributions are:
\begin{itemize}
\item A physics-simulation fluid video dataset spanning pouring and sloshing behaviors under systematically varied physical parameters, paired with a curated real-video benchmark, together comprising $3{,}958$ training videos and two held-out test sets.
\item A dual-stream video diffusion architecture that adds an explicit, zero-convolution-fused optical-flow decoding branch to a frozen pretrained backbone, trained with a joint RGB--flow objective, at a cost of only $\approx\!2\%$ additional decoder parameters.
\item Extensive quantitative and qualitative evaluation across two model scales (1.3B/14B parameters), two test distributions, a state-of-the-art competitor, a $36$-participant human study, and a direct optical-flow accuracy probe, all of which show consistent gains in physical plausibility without degrading text alignment.
\end{itemize}

\section{Related Work}
\label{sec:related}

\paragraph{Video diffusion generation.}
Diffusion-based video generators~\cite{videodiffusion2022,makeavideo2022,cogvideox2024,wan2025} have progressively scaled from short, low-resolution clips to minute-scale, high-definition, text- and image-conditioned generation, typically by operating a diffusion transformer over a compressed latent produced by a video VAE. Beyond raw generation capacity, systematic alignment frameworks have recently been proposed to improve the fidelity, controllability, and robustness of such models~\cite{liang2026teleboost}. Our method is architecture-agnostic and is instantiated on top of such a pretrained image-to-video diffusion transformer, which we treat as a frozen backbone.

\paragraph{Physically-grounded video and 4D generation.}
A growing line of work injects explicit physics into generative pipelines: rigid-body physics engines drive image-to-video synthesis~\cite{physgen2024}; physics-integrated 3D Gaussians couple a differentiable simulator with a Gaussian-splatting renderer for object dynamics~\cite{physgaussian2024}; and video-diffusion priors are distilled to recover or regularize physical material parameters for 3D assets~\cite{physdreamer2024,dreamphysics2025,phys4dgen2025}. These methods largely target rigid or piecewise-rigid object dynamics with a known 3D representation. In contrast, our target ---pouring and sloshing fluids--- involves continuous topology change that is difficult to represent with a fixed mesh or particle skeleton, so we instead inject physical supervision indirectly, through simulation-generated \emph{training data} and an explicit \emph{motion-field} objective, rather than through a differentiable simulator in the generation loop. Situating this effort within a broader agenda, a recent roadmap argues for coupling physical simulation with world-action models as a path toward embodied physical intelligence~\cite{liang2026world}, while a concurrent survey charts how reinforcement-learning techniques are being integrated with visual generative models to improve controllability and physical fidelity~\cite{liang2026integrating}; our simulation-data-plus-explicit-motion-objective design can be viewed as one concrete instantiation of this broader direction, specialized to fluid dynamics.

\paragraph{Optical flow as a training signal.}
Optical flow has long served as an intermediate representation for motion understanding; RAFT~\cite{raft2020} established an accurate, efficient recurrent flow estimator that we use both to construct pseudo-ground-truth flow supervision for training and as an independent evaluation tool for measuring the motion self-consistency of generated videos. Unlike prior work that uses optical flow purely as a post-hoc evaluation or warping mechanism, we introduce flow prediction as an auxiliary decoding branch trained jointly with RGB reconstruction, so that the motion objective directly shapes the shared video latent representation during generation.

\paragraph{Simulation-generated training data.}
Particle-based simulation frameworks that couple 3D Gaussian Splatting with physical solvers~\cite{physgaussian2024,gaussiansplashing2025,versagauss2025} enable photorealistic rendering of physically simulated dynamics from a single input image, providing a practical route to large-scale, parameter-controlled synthetic video generation. We build our fluid dataset on such a multiphase particle simulation framework~\cite{versagauss2025}, using it purely as a \emph{data generation} tool that is decoupled from the downstream video generation model.

\section{Method}
\label{sec:method}

\subsection{Physics-Simulation Fluid Dataset}
\label{sec:dataset}

Because no existing corpus pairs fluid video with dense, physically grounded motion parameters at scale, we construct a dedicated training and evaluation dataset spanning two canonical fluid behaviors.

\begin{figure}[t]
\centering
\includegraphics[width=\linewidth]{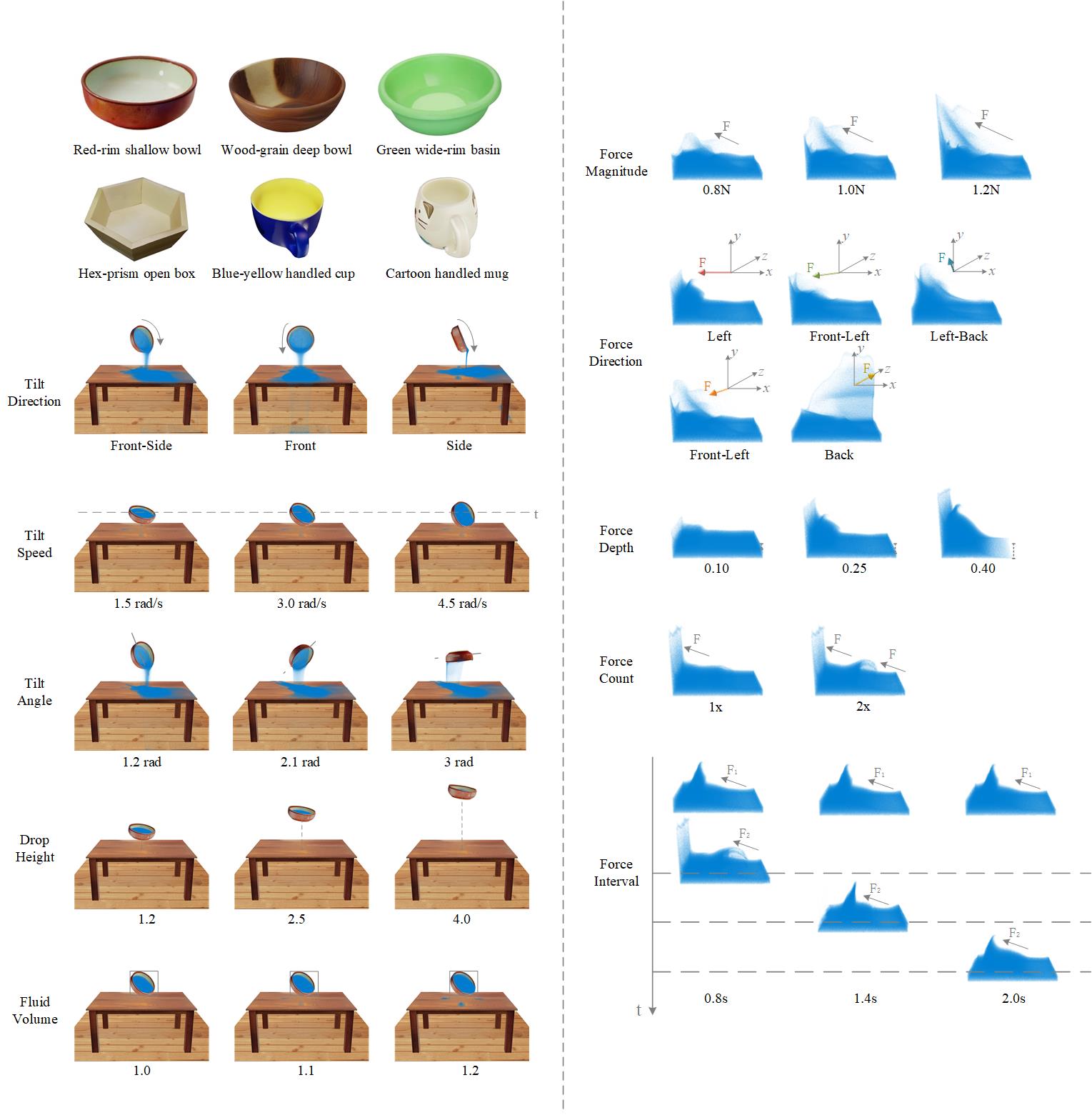}
\caption{\textbf{Dataset overview.} Left: pouring scenario with varied parameters including container geometries, tilt direction/speed/angle, drop height, and fluid volume. Right: sloshing scenario with force parameters including magnitude, direction, depth, force count, and interval. Renders and simulations are generated using the multiphase particle simulator of~\cite{versagauss2025}.}
\label{fig:containers}
\end{figure}

\paragraph{Pouring scenario.}
We simulate a container-tipping scene containing four physical entities --- a container, a table, the floor, and the fluid itself, modeled with a weakly-compressible constitutive law --- rendered using the multiphase particle-simulation framework of~\cite{versagauss2025}. The container's Gaussian ellipsoids are rigidly frozen relative to one another to avoid unwanted deformation during rotation. We designed six containers spanning distinct geometric families (Fig.~\ref{fig:containers}): a shallow ceramic bowl, a deep wood-grain bowl, a wide-brim plastic basin, a hexagonal open box, a handled cup, and a cartoon mug, so that the model is exposed to varied rim geometry and pouring affordances. For each container we systematically vary five physically meaningful parameters: tilt direction (front-side, front, side; 3 values), tilt angular speed ($1.5$, $3.0$, $4.5\,\mathrm{rad/s}$), final tilt angle ($1.2$, $2.1$, $3.0\,\mathrm{rad}$), container-to-table drop height (three simulation-unit levels), and fluid volume (three relative levels). The full combinatorial sweep ($6\times3\times3\times3\times3\times3$) yields $1{,}458$ pouring videos, each $7.5\,$s long at $150$ frames, capturing the complete process from the onset of tipping to the fluid settling on the ground.

\paragraph{Sloshing scenario.}
To capture fluid response to external perturbation without occluding containers, we simplify the scene to the fluid volume alone, bounded by a fixed rectangular boundary, and apply a horizontal impulse force for a controlled duration. We vary force magnitude ($0.8$, $1.0$, $1.2\,\mathrm{N}$), force direction (5 in-plane directions), force application depth (3 levels), and application pattern (a single impulse, or a repeated impulse train at one of 3 inter-impulse intervals of $0.8/1.4/2.0\,$s). The sweep ($3\times5\times3\times(1+1\times3)$) produces $180$ sloshing videos, each $25\,$s long at $500$ frames, recording the fluid's transition from rest through sloshing back to equilibrium.

\paragraph{Real video collection.}
To retain exposure to real-world visual statistics, we mine a large stock-footage corpus ($10{,}000$ professionally shot videos with automatically generated short/long captions) for fluid-related content: we keep clips whose short caption jointly contains a fluid noun (\emph{water}, \emph{liquid}, \emph{fluid}, \emph{etc.}) and a pouring verb (\emph{dump}, \emph{spill}, \emph{tip}, \emph{pour}, \emph{etc.}), yielding $3{,}835$ curated real fluid videos.

\begin{figure}[t]
\centering
\includegraphics[width=\linewidth]{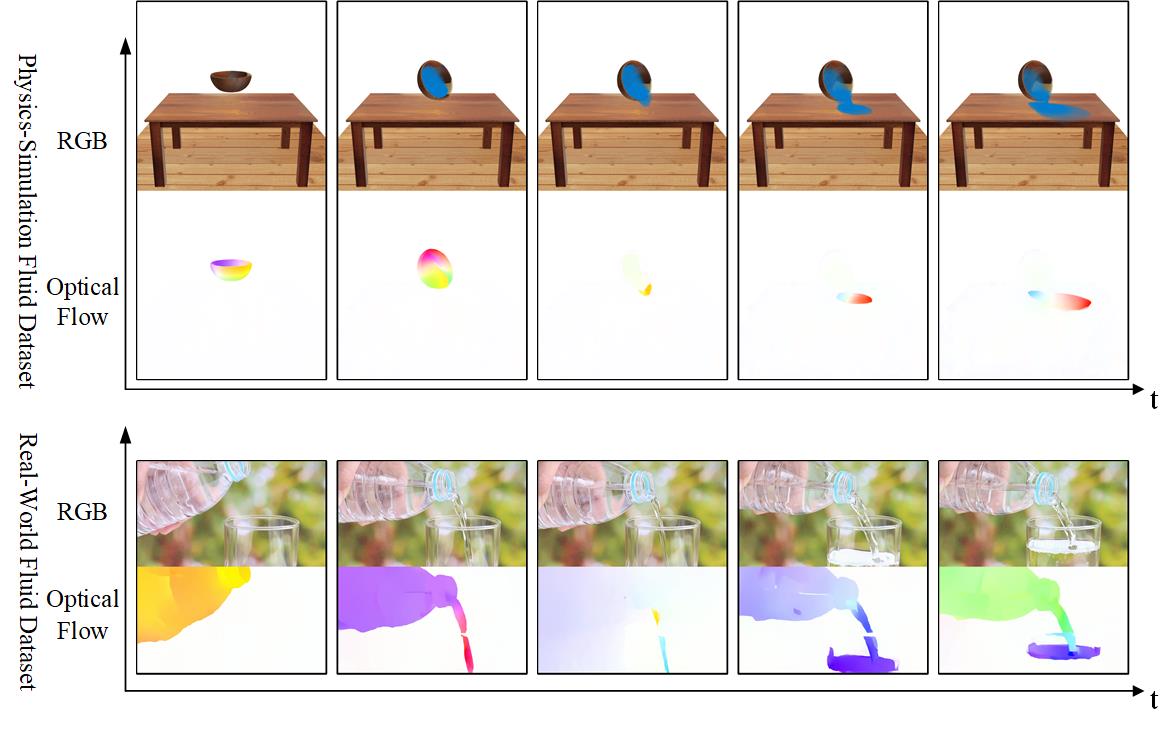}
\caption{\textbf{RGB and pseudo-ground-truth optical-flow supervision.} Representative frame sequences from the simulated (top) and real (bottom) training data, each with the corresponding RAFT-extracted, foreground-masked flow field used as pseudo ground truth for $\mathcal L_f$ (Eq.~\ref{eq:flow}).}
\label{fig:flowsamples}
\end{figure}

\paragraph{Preprocessing and pseudo-flow supervision.}
All videos are unified to $81$ frames. For simulated videos we crop the temporal window that contains the principal motion (frames $34$--$115$ for pouring, $179$--$260$ for sloshing); for real videos, whose motion onset is unknown and whose duration varies, we sample $81$ frames uniformly. We then extract dense optical flow with a pretrained RAFT model~\cite{raft2020} for every training video to serve as pseudo ground-truth motion supervision. Because the simulated videos are rendered against a texture-less white background, naive flow extraction spuriously assigns motion to the static background; we suppress this by masking the extracted flow with the simulator's foreground segmentation, zeroing the flow outside the fluid/container region so that the supervision signal is concentrated on genuine fluid motion (Fig.~\ref{fig:flowsamples}). The final training set combines $1{,}638$ simulated videos with $2{,}320$ real videos ($3{,}958$ total); the remaining $1{,}515$ real videos form our primary test set, which we refer to as \textbf{iStock-Fluid}. We additionally construct a generalization benchmark, \textbf{Flux-Fluid}, by writing $18$ fluid-related text prompts and generating their first frames with a text-to-image model~\cite{flux2025}, decoupling evaluation from any visual bias present in the real-video corpus.

\subsection{Dual-Stream Video Generation Architecture}
\label{sec:arch}

We target the image-to-video (I2V) setting: given a first frame and a text prompt, generate a temporally coherent video. Our backbone is a pretrained diffusion-transformer I2V model~\cite{wan2025} composed of a video encoder (\textit{WanEncoder}), a stack of $N$ diffusion-transformer (DiT) blocks conditioned on a text encoder, and an RGB decoder. Standard fine-tuning of such a backbone optimizes only a photometric reconstruction objective, which we hypothesize under-constrains motion: the model can minimize per-frame RGB error while reproducing an appearance pattern memorized from limited training footage, without learning a generalizable notion of pixel-level motion.

\begin{figure*}[t]
\centering
\includegraphics[width=\linewidth]{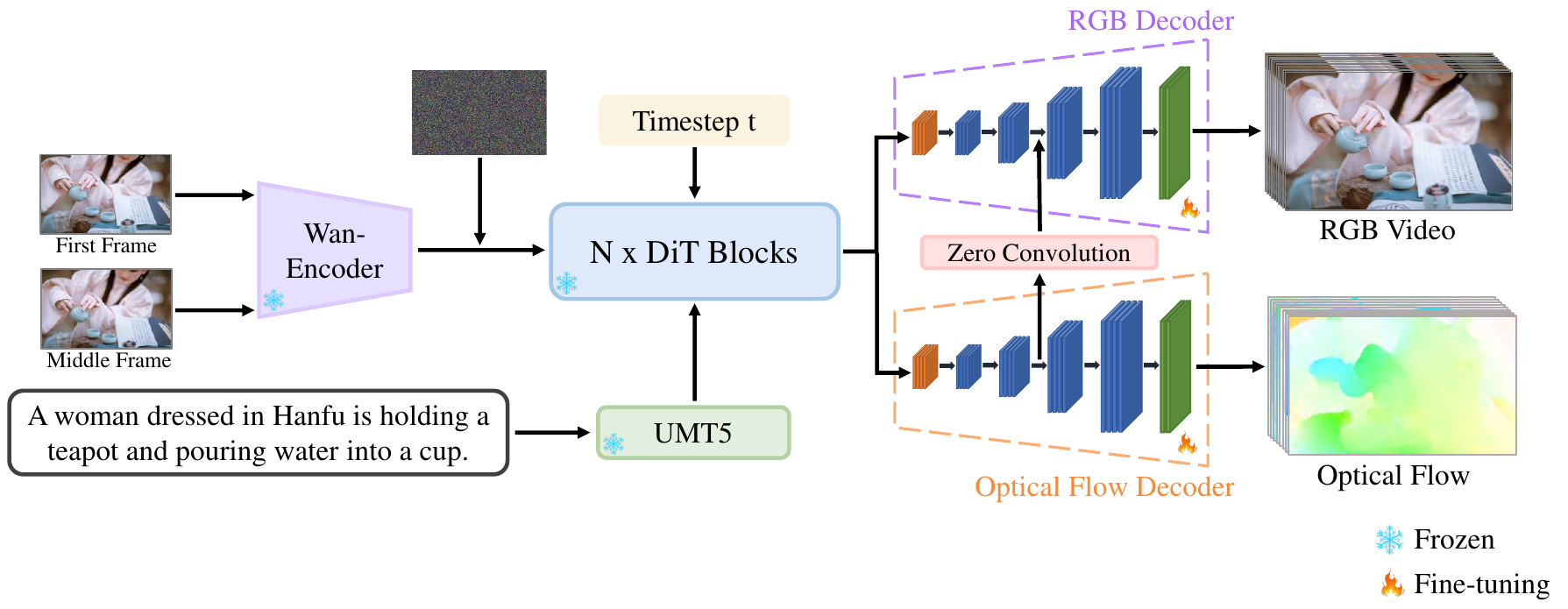}
\caption{\textbf{Dual-stream architecture.} A frozen WanEncoder and stack of DiT blocks (conditioned on a text prompt via a frozen text encoder) produce a shared latent that is decoded by two branches: a fine-tuned RGB Decoder and a new, lightweight Optical-Flow Decoder. Flow-decoder features are injected into the RGB stream through zero-initialized convolutions, so the flow branch has no effect at the start of training and is gradually blended in as its zero-convolution weights are learned.}
\label{fig:arch}
\end{figure*}

We therefore introduce a second decoding branch, the \textbf{Optical-Flow Decoder}, that shares the RGB decoder's architectural skeleton (a 3D input convolution, multi-scale upsampling stages, and a prediction head) but is configured with narrower channel widths, so that its parameter count is only $\approx\!2\%$ of the RGB decoder's. Both decoders consume the same DiT output latent $x_{\text{latent}}$ at each diffusion step and independently produce an RGB video $\hat v$ and a per-pixel flow field $\hat f$ (Fig.~\ref{fig:arch}).

\paragraph{Zero-convolution fusion.}
To let the flow branch inform RGB generation without destabilizing the pretrained decoder, each intermediate feature map produced by the flow decoder is passed through an independent zero-initialized convolution and added element-wise to the corresponding intermediate feature map in the RGB decoder. Because the fusion convolution starts at zero, the flow branch has no effect on RGB generation at the start of fine-tuning, guaranteeing that initial output quality matches the pretrained backbone; as training proceeds, the fusion weights are learned, progressively injecting motion-aware features into the RGB stream. This produces a mutually reinforcing training dynamic: the RGB decoder receives motion-aware guidance from the flow branch, while the flow branch receives sharper geometric boundary cues from the jointly optimized RGB reconstruction.

\paragraph{Conditioning.}
In addition to the first frame, we condition on an intermediate frame (the $10$th frame in our experiments), selected via a binary mask mechanism: the first and intermediate frames are VAE-encoded and concatenated with the noise latent along the channel dimension, with an accompanying binary mask flagging them as conditioning frames (mask value $1$) versus frames to be generated (mask value $0$). The first frame anchors the static scene layout and container pose, while the intermediate frame carries an early cue about the container's rotational speed and direction, steering the model to focus capacity on the fluid's own dynamics rather than re-deriving rigid-body container motion from scratch.

\paragraph{Partial freezing.}
We freeze \textit{WanEncoder}, all DiT blocks, and the text encoder, updating only the RGB decoder and the new Optical-Flow decoder. The frozen components already encode broad visual and temporal priors learned from large-scale pretraining; freezing them mitigates overfitting to our comparatively small fluid dataset, reduces the trainable parameter count, and concentrates learning capacity on the decoder-level mapping from shared latents to physically consistent motion and appearance.

\subsection{Joint RGB--Flow Objective}
\label{sec:loss}

We train the two decoders with a joint loss
\begin{equation}
\mathcal L = \mathcal L_v + \mathcal L_f,
\label{eq:total}
\end{equation}
where $\mathcal L_v$ supervises RGB reconstruction and $\mathcal L_f$ supervises the predicted flow field.

The RGB loss combines a per-pixel reconstruction term with a latent-space distribution matching term,
\begin{equation}
\mathcal L_v = \mathcal L_{\mathrm{MAE}}(v,\hat v) + \lambda_{\mathrm{kl}}\, \mathcal L_{\mathrm{kl}}(v,\hat v), \quad \lambda_{\mathrm{kl}} = 0.1,
\label{eq:rgb}
\end{equation}
where $v$ and $\hat v$ denote the ground-truth and predicted frames, $\mathcal L_{\mathrm{MAE}}$ is an outlier-robust mean-absolute-error reconstruction loss, and $\mathcal L_{\mathrm{kl}}$ regularizes the predicted latent distribution towards the ground-truth latent distribution.

The flow loss combines an end-point-error term with two regularizers,
\begin{equation}
\mathcal L_f = \mathcal L_{\mathrm{epe}}(f,\hat f) + \lambda_{\mathrm{ss}}\, \mathcal L_{\mathrm{ss}}(f,\hat f) + \lambda_{\mathrm{ts}}\, \mathcal L_{\mathrm{ts}}(f,\hat f),
\label{eq:flow}
\end{equation}
with $\lambda_{\mathrm{ss}} = 0.01$ and $\lambda_{\mathrm{ts}} = 0.05$. $\mathcal L_{\mathrm{epe}}$ is the standard per-pixel Euclidean distance between predicted and pseudo-ground-truth flow, directly penalizing displacement error. $\mathcal L_{\mathrm{ss}}$ is an edge-aware spatial smoothness term that encourages flow continuity within homogeneous image regions while permitting discontinuities at object boundaries such as the fluid--container interface, matching the physical expectation that fluid motion is smooth away from boundaries but can change abruptly at them. $\mathcal L_{\mathrm{ts}}$ penalizes frame-to-frame flow variation, discouraging jitter and enforcing that the generated motion evolves in a temporally coherent, physically plausible manner.

By jointly minimizing Eq.~(\ref{eq:total}), the model is required not only to produce visually convincing RGB frames but also to internalize a self-consistent, per-pixel motion field, coupling appearance generation to an explicit physical-motion objective.

\subsection{Training Procedure}
\label{sec:training}

Algorithm~\ref{alg:train} summarizes the training loop. Only the decoder parameters $\Theta_{\mathrm{dec}}$ (RGB Decoder and Optical-Flow Decoder) are updated with AdamW at an initial learning rate of $1\times10^{-5}$ under a cosine annealing schedule, following the timestep-dependent loss weighting scheme of the base model~\cite{wan2025}. Training videos are resized to $512\times512$ resolution, $81$ frames, $16$ fps. We fine-tune two backbone scales, $1.3$B and $14$B parameters, to study whether our gains persist as model capacity grows; the $1.3$B model is trained with a per-GPU batch size of $1$ on $8$ GPUs and the $14$B model on $16$ GPUs, both for $20$ epochs, using ZeRO-2 distributed optimization.

\begin{algorithm}[t]
\caption{Dual-Stream Training}
\label{alg:train}
\begin{algorithmic}[1]
\Require Pretrained backbone $\Theta_{\mathrm{Wan}}$, trainable decoder params $\Theta_{\mathrm{dec}}$, dataset $\mathcal D$, AdamW, learning rate $\eta$
\State Freeze WanEncoder, DiT, text encoder; set only $\Theta_{\mathrm{dec}}$ trainable
\For{each epoch}
  \For{each batch $(v_{\mathrm{rgb}}, f_{\mathrm{gt}}, \mathrm{prompt}) \in \mathcal D$}
    \State $z \gets \mathcal E_{\mathrm{VAE}}(v_{\mathrm{rgb}})$ \Comment{encode to latent}
    \State sample timestep $t$; $z_t \gets \mathrm{add\_noise}(z,t)$
    \State $\hat z_0 \gets \mathrm{denoise}(z_t, \mathrm{prompt})$ via frozen DiT
    \State $\hat f \gets \mathcal D_{\mathrm{flow}}(\hat z_0)$; \quad $\hat v_{\mathrm{rgb}} \gets \mathcal D_{\mathrm{rgb}}(\hat z_0)$
    \State compute $\mathcal L_v$ (Eq.~\ref{eq:rgb}), $\mathcal L_f$ (Eq.~\ref{eq:flow})
    \State $\mathcal L \gets w(t)\cdot(\mathcal L_v + \mathcal L_f)$
    \State $\Theta_{\mathrm{dec}} \gets \Theta_{\mathrm{dec}} - \eta \nabla_{\Theta_{\mathrm{dec}}} \mathcal L$
  \EndFor
\EndFor
\end{algorithmic}
\end{algorithm}

\section{Experiments}
\label{sec:exp}

\subsection{Setup}
We train on the $3{,}958$-video mixed dataset of Sec.~\ref{sec:dataset} and evaluate on iStock-Fluid ($1{,}515$ real videos) and Flux-Fluid ($18$ text prompts with Flux-generated first frames). All videos are unified to $81$ frames, $512\times512$, $16$ fps. At inference, the flow decoder adds only $\approx\!2\%$ parameters relative to the RGB decoder, so overall compute is close to the unmodified backbone; generating a $5$-second, $480$p ($854\times480$) video takes $\approx\!4$ minutes on a single RTX~4090.

\paragraph{Metrics.} Following the VIDEOPHY-2 protocol~\cite{videophy2_2025}, we report \textbf{Physical Commonsense (PC)}, the percentage of generated videos scoring $\geq\!4$ on a $5$-point automated physical-plausibility rating (gravity-consistent falling, plausible container collisions, mass-conserving level rise, splash direction consistent with impact force), and \textbf{Video Quality (VQ)}, the percentage of videos scoring $\geq\!4$ on \emph{both} PC and Semantic Adherence (SA, prompt--content alignment), capturing joint physical and semantic fidelity. We additionally evaluate flow prediction accuracy directly, using RAFT-extracted flow from the generated video as a reference and comparing it to the flow decoder's own output via End-Point-Error (EPE), the fraction of pixels with error $>\!1$px and $>\!3$px, and F1-all (fraction with error $>\!3$px \emph{and} relative error $>\!5\%$).

\subsection{Quantitative Comparison}
Table~\ref{tab:main} compares our dual-stream fine-tuned models against the unmodified backbone at both scales. On iStock-Fluid, our $1.3$B model improves PC from $66.92$ to $70.77$ and VQ from $56.92$ to $60.00$; the $14$B model improves PC from $70.00$ to $78.75$ and VQ from $57.50$ to $61.25$. Gains persist under the visually distinct Flux-Fluid distribution (PC $70.59\!\rightarrow\!76.47$, VQ $70.59\!\rightarrow\!76.47$ for the $1.3$B model), indicating that the improvement generalizes beyond the real-video training distribution rather than merely overfitting to it, and that it holds across a $10\times$ change in backbone scale.

\begin{table}[t]
\centering
\small
\caption{\textbf{Backbone comparison.} PC/VQ ($\%$, higher is better) on iStock-Fluid and Flux-Fluid, before and after our dual-stream fine-tuning, at two model scales.}
\label{tab:main}
\begin{tabular}{@{}llcc@{}}
\toprule
Test set & Model & PC$\uparrow$ & VQ$\uparrow$ \\
\midrule
\multirow{4}{*}{iStock-Fluid}
 & Wan2.1-1.3B & 66.92 & 56.92 \\
 & \textbf{Ours-1.3B} & \textbf{70.77} & \textbf{60.00} \\
 & Wan2.1-14B & 70.00 & 57.50 \\
 & \textbf{Ours-14B} & \textbf{78.75} & \textbf{61.25} \\
\midrule
\multirow{2}{*}{Flux-Fluid}
 & Wan2.1-1.3B & 70.59 & 70.59 \\
 & \textbf{Ours-1.3B} & \textbf{76.47} & \textbf{76.47} \\
\bottomrule
\end{tabular}
\end{table}

Table~\ref{tab:sota} additionally compares against a leading open-source competitor, CogVideoX-5B~\cite{cogvideox2024}, on a $51$-video sample of iStock-Fluid. Our method achieves the best PC ($66.67$) and VQ ($60.78$) among the three, ahead of both Wan2.1 and CogVideoX-5B, consistent with the qualitative findings in Sec.~\ref{sec:qual}.

\begin{table}[t]
\centering
\small
\caption{\textbf{Comparison with a state-of-the-art competitor} on a $51$-video iStock-Fluid sample.}
\label{tab:sota}
\begin{tabular}{@{}lcc@{}}
\toprule
Model & PC$\uparrow$ & VQ$\uparrow$ \\
\midrule
CogVideoX-5B~\cite{cogvideox2024} & 62.75 & 52.94 \\
Wan2.1~\cite{wan2025} & 62.75 & 58.82 \\
\textbf{Ours} & \textbf{66.67} & \textbf{60.78} \\
\bottomrule
\end{tabular}
\end{table}

Because Ours and Wan2.1 share an identical backbone and differ only in the addition of the flow-decoder branch and fluid-focused fine-tuning, Tables~\ref{tab:main}--\ref{tab:sota} also serve as an internal ablation: the consistent PC/VQ improvement over the frozen baseline, at both the $1.3$B and $14$B scales and under two different test distributions, isolates the benefit of the proposed dual-stream, flow-supervised fine-tuning procedure.

\subsection{Qualitative Comparison}
\label{sec:qual}
\begin{figure}[t]
\centering
\includegraphics[width=\linewidth]{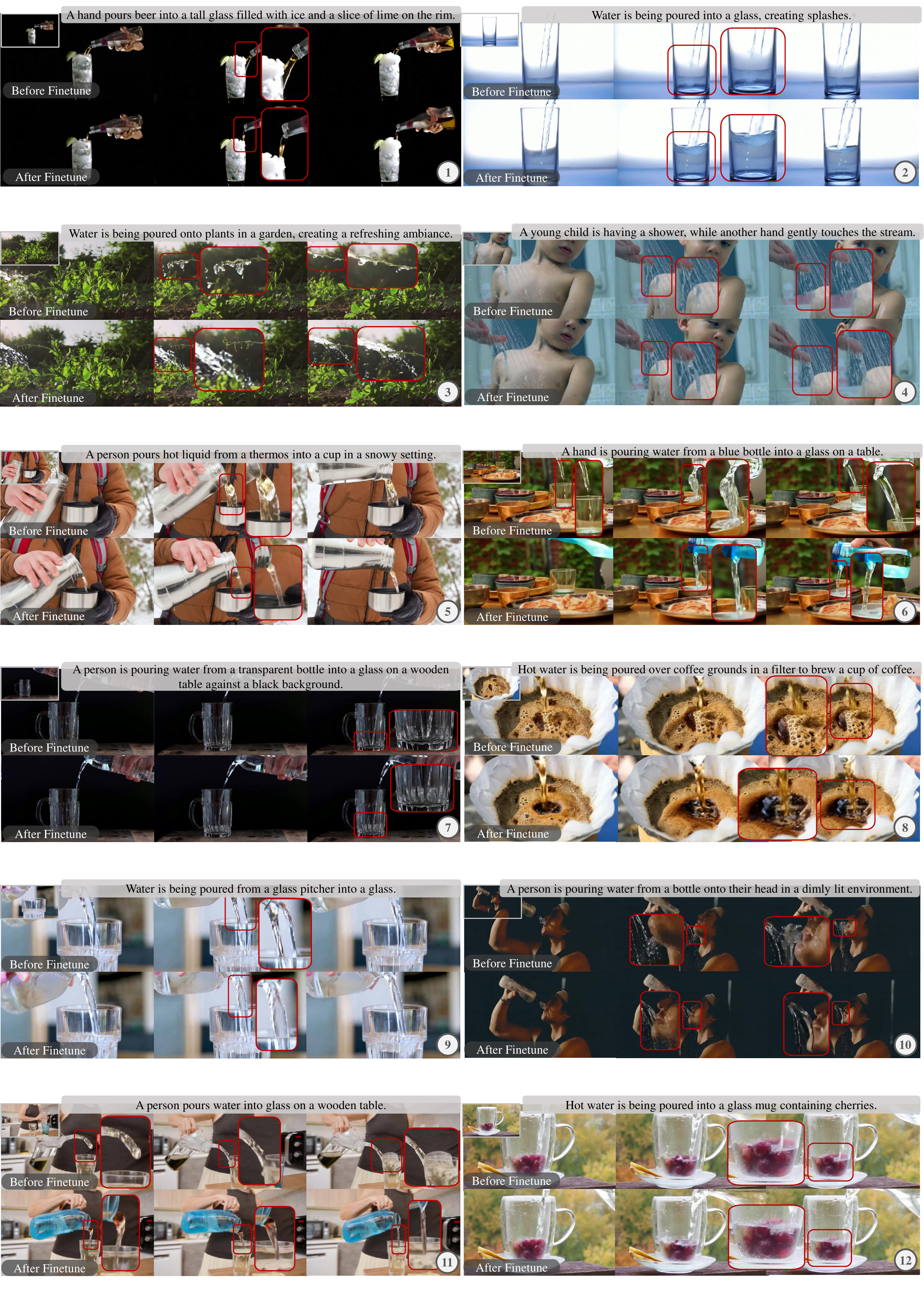}
\caption{\textbf{Qualitative comparison (1.3B).} Our fine-tuned model (bottom row of each pair) vs.\ the unmodified Wan2.1-1.3B backbone (top row), across pouring, watering, and shower scenes. Our model produces rim-consistent stream entry, smooth liquid-level rise, and momentum-consistent splash/foam interaction.}
\label{fig:qual}
\end{figure}

\begin{figure}[t]
\centering
\includegraphics[width=\linewidth]{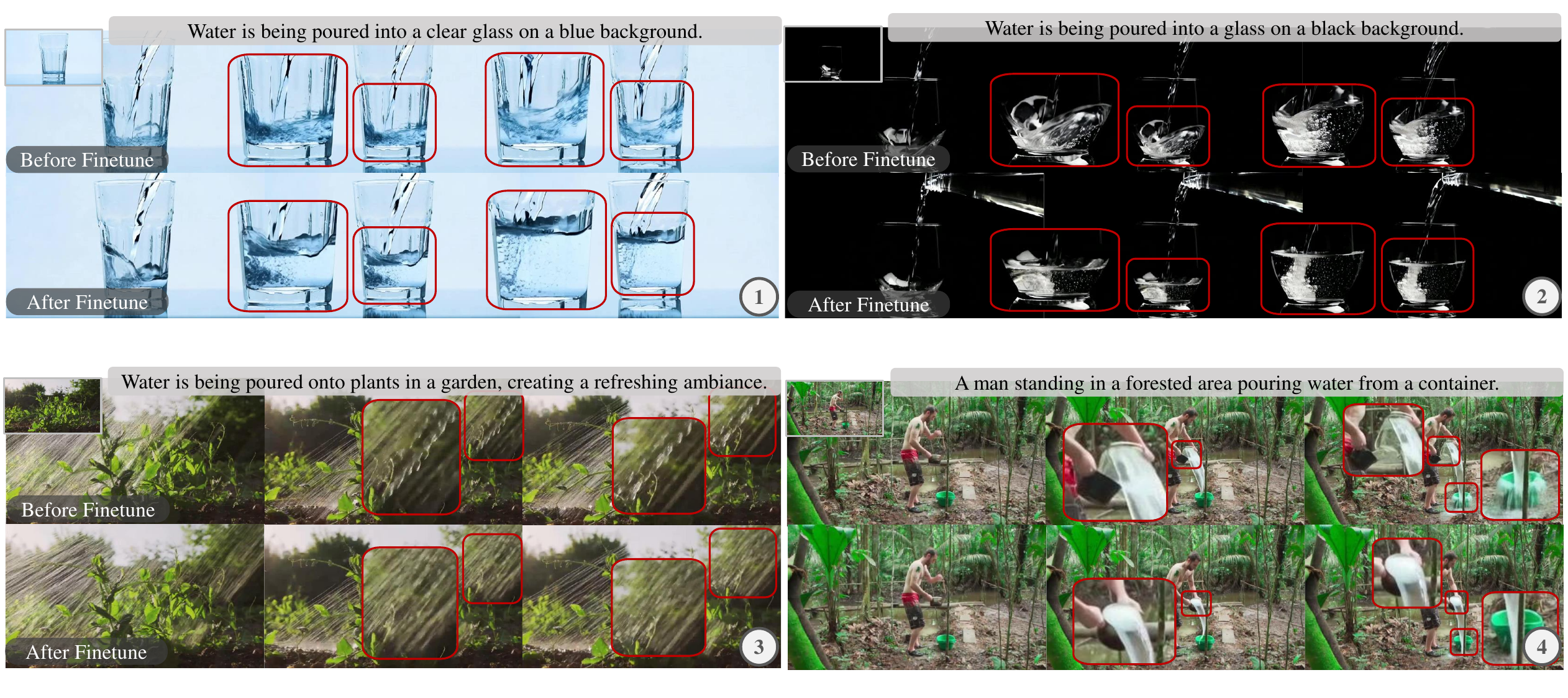}
\caption{\textbf{Qualitative comparison (14B).} Same comparison as Fig.~\ref{fig:qual} at the larger 14B backbone scale.}
\label{fig:qual14b}
\end{figure}

\begin{figure}[t]
\centering
\includegraphics[width=\linewidth]{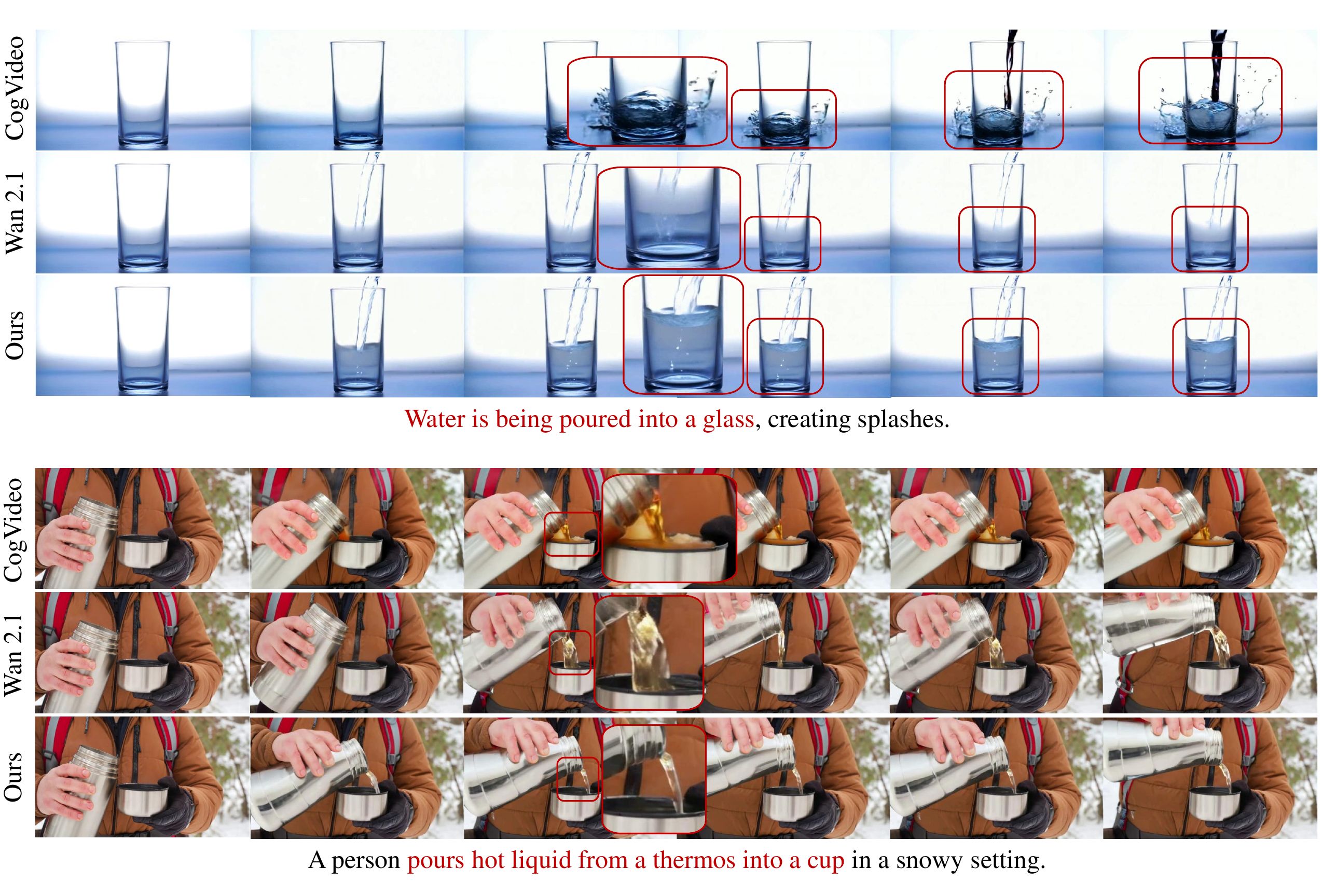}
\caption{\textbf{Comparison with Wan2.1 and CogVideoX-5B~\cite{cogvideox2024}} on two pouring prompts; our method most consistently respects rim-of-entry, level rise, and splash-direction constraints.}
\label{fig:qual_sota}
\end{figure}

Figure~\ref{fig:qual} contrasts our $1.3$B model against the unmodified Wan2.1-1.3B backbone across representative pouring, watering, and shower scenes. The unmodified backbone exhibits two recurring failure patterns: (i) unstable fluid geometry, including column breakage, sudden volume change, and gravity-inconsistent bouncing or floating, reflecting insufficient modeling of continuity and momentum conservation; and (ii) inconsistent fluid--container/environment interaction, including offset entry points, discontinuous liquid-level changes, and splash directions misaligned with the applied force. Our fine-tuned model instead produces streams that enter consistently at the container rim and follow a gravity-consistent trajectory, liquid levels that rise smoothly with the poured volume, and splash/foam interactions --- e.g.\ a poured stream visibly displacing existing foam in a coffee cup --- that better respect momentum transfer at the fluid--fluid interface. Figure~\ref{fig:qual14b} repeats the comparison at the $14$B scale: even though the larger backbone has stronger baseline generation quality, it exhibits the same qualitative failure modes (non-rising liquid levels, unrealistic hinge-like water trajectories when watering plants), which our fine-tuning likewise corrects. Figure~\ref{fig:qual_sota} extends the comparison to CogVideoX-5B~\cite{cogvideox2024}, whose splashes frequently disperse in an unconstrained, momentum-inconsistent pattern; our method again produces the most physically coherent trajectory among the three methods compared.

\subsection{Human Study}
To complement the automated PC/VQ metrics, we ran a blind human study with $36$ participants across a range of ages and occupations. For $10$ fluid scenes, each participant rated $3$ blindly ordered, method-anonymized clips (Ours, Wan2.1-1.3B, CogVideoX-5B) on a $4$-point realism scale ($1$ = least, $4$ = most realistic), for $30$ ratings per participant. Table~\ref{tab:user} reports the mean and standard deviation across all participants and scenes. The ranking mirrors the automated metrics: our method is rated most realistic on average, roughly $0.6$ points above Wan2.1 on the $4$-point scale, indicating that the physical-plausibility gains captured by PC/VQ are also perceptible to human observers.

\begin{table}[t]
\centering
\small
\caption{\textbf{Human study.} Mean realism rating (4-point scale, higher is better) with standard deviation, $36$ raters $\times$ $10$ scenes.}
\label{tab:user}
\begin{tabular}{@{}lcc@{}}
\toprule
Model & Mean rating$\uparrow$ & Std.\ dev. \\
\midrule
CogVideoX-5B~\cite{cogvideox2024} & 1.2500 & 0.5000 \\
Wan2.1~\cite{wan2025} & 2.2500 & 1.0522 \\
\textbf{Ours} & \textbf{2.8667} & 1.3074 \\
\bottomrule
\end{tabular}
\end{table}

\subsection{Optical Flow Accuracy}
Finally, we directly probe whether the model has learned a physically coherent motion field, independent of the RGB-based PC/VQ metrics. We extract RAFT flow from our model's own \emph{generated} video and compare it, as a reference, to the flow predicted by the Optical-Flow Decoder for the same sample (Table~\ref{tab:flow}). On iStock-Fluid, the two agree to an EPE of only $0.538$ pixels, with $0\%$ of pixels exceeding a $3$-pixel error and $14.7\%$ exceeding $1$ pixel (concentrated at high-frequency fluid--container boundaries). On the more distributionally distant Flux-Fluid set, EPE rises to $1.541$ pixels, with $21.8\%$ and $10.0\%$ of pixels exceeding $1$ and $3$ pixels respectively, still within an acceptable range given the domain shift. These results indicate that the explicit flow objective yields a self-consistent, generalizable internal motion representation, rather than a flow branch that is decoupled from what the RGB stream actually renders.

\begin{table}[t]
\centering
\small
\caption{\textbf{Optical-flow self-consistency.} Error between the decoder's predicted flow and RAFT flow re-extracted from the generated video.}
\label{tab:flow}
\begin{tabular}{@{}lcccc@{}}
\toprule
Test set & EPE$\downarrow$ & $>\!1$px$\downarrow$ & $>\!3$px$\downarrow$ & F1-all$\downarrow$ \\
\midrule
iStock-Fluid & 0.538 & 14.7\% & 0.0\% & 4.680 \\
Flux-Fluid & 1.541 & 21.8\% & 10.0\% & 10.040 \\
\bottomrule
\end{tabular}
\end{table}

\section{Future Work}
\label{sec:limit}
Our dataset currently spans two canonical fluid behaviors, pouring and sloshing; a natural extension is to broaden the underlying multiphase simulator~\cite{versagauss2025} to additional scenarios such as phase change (boiling, freezing) and fluid--fluid mixing across distinct viscosities, further widening the range of physical motion the model is exposed to. The simulated videos are rendered against a simplified background, and foreground flow-masking already concentrates supervision on genuine fluid motion; extending pseudo-ground-truth flow supervision to weakly-labeled internet video is a promising way to further narrow the residual simulation-to-real gap reflected in the higher Flux-Fluid EPE relative to iStock-Fluid.

Our physics-grounded fine-tuning is complementary to, and can be directly built upon by, reinforcement-learning-style post-training. Because the PC/VQ physical-plausibility score used for evaluation in Sec.~\ref{sec:exp} already yields a scalar signal for any generated clip, it is well suited to serve as a reward for a post-training stage applied on top of our fine-tuned model. Recent post-training and alignment techniques for visual generation offer concrete recipes for such a stage: self-paced curricula that adapt sampling difficulty to the generator's current competence~\cite{li2025growing}, preference-based policy optimization~\cite{ni2026seeing}, objective-aware credit assignment along the denoising trajectory~\cite{li2026learning}, reward-aware trajectory shaping for few-step samplers~\cite{li2026reward}, and Bayesian prior-guided optimization~\cite{liu2026learning}. Chaining one of these procedures, driven by our PC/VQ scorer, onto our physics-grounded fine-tuning is a natural next step towards further strengthening the physical plausibility gains reported in this paper.

\section{Conclusion}
\label{sec:conclusion}
We presented a physics-simulation approach to improving the physical realism of fluid video generation. We constructed a controlled fluid dataset spanning pouring and sloshing behaviors, combined it with curated real footage, and used it to train a dual-stream fine-tuning architecture that augments a frozen pretrained video diffusion transformer with an explicit, zero-convolution-fused optical-flow decoding branch. Across two model scales, two test distributions, a state-of-the-art competitor, a blind human study, and a direct flow-accuracy probe, our method consistently improves physical plausibility while preserving the generation quality and generalization ability of the underlying pretrained backbone.

{\small
\bibliographystyle{plain}

}

\end{document}